\documentclass[acmsmall, authorversion, nonacm]{acmart}

\usepackage{booktabs} 

\usepackage{amsfonts,amsmath}
\usepackage{algpseudocode}
\usepackage{algorithmicx}
\usepackage{graphicx,epsfig}
\usepackage{color}
\usepackage[ruled]{algorithm2e} 

\newcommand {\mymarginpar}[1]{\marginpar{#1}}
\renewcommand {\marginpar}[1]{}

\def\_{\rule{.3em}{.15ex}}      

\newcommand{\ls}[1]
   {\dimen0=\fontdimen6\the\font
    \lineskip=#1\dimen0
    \advance\lineskip.5\fontdimen5\the\font
    \advance\lineskip-\dimen0
    \lineskiplimit=.9\lineskip
    \baselineskip=\lineskip
    \advance\baselineskip\dimen0
    \normallineskip\lineskip
    \normallineskiplimit\lineskiplimit
    \normalbaselineskip\baselineskip
    \ignorespaces
   }

\newcommand {\bearn}{\begin{eqnarray*}}
\newcommand {\eearn}{\end{eqnarray*}}
\newcommand {\barr}{\begin{array}}
\newcommand {\earr}{\end{array}}

\newcommand {\benum} {\begin{enumerate}}
\newcommand {\eenum} {\end{enumerate}}

\newcommand {\bdesc} {\begin{description}}
\newcommand {\edesc} {\end{description}}

\newcommand {\bfig}[2] {\begin{figure}
  \centering
  \includegraphics[width=#2]{#1}}
\newcommand {\brotatefig}[2] {\begin{figure}[htbp]
                        \centerline {
                         \epsfig{figure={#1},clip=,angle=-90,width={#2}}}}
\newcommand {\bfigfirst}[2] {\begin{figure}[h]
                        \centerline {
                        \setlength{\epsfxsize}{#2}
                        \epsffile{#1}}}
\newcommand {\efig}[2]{ \caption{#2}
                        \label{fig:#1}
                        \end{figure}
                        \mymarginpar{fig:#1}}
\newcommand {\erotatefig}[2]{ \caption{#2}
                        \label{fig:#1}
                        \end{figure}
                        \mymarginpar{fig:#1}}
\newcommand {\rfig}[1]{Figure \ref{fig:#1}}

\newcommand {\btab}[1]{
                       \begin{table}
                       \centering
                       \begin{tabular}{#1}}
\newcommand {\etab}[3] {
                       \end{tabular}
                       \caption[#3]{#2}
                       \label{tab:#1}
                       \end{table}
                       \mymarginpar{tab:#1}
                       \vspace{.1in}}

\newcommand {\btabular}[1]{\begin{center}
                       \begin{tabular}{#1}}
\newcommand {\etabular}{\end{tabular}
                       \end{center}}

\newcommand {\bdefin}[1]{\begin{definition}
                      \mymarginpar{def:#1}
                      \label{def:#1} }
\newcommand {\edefin}       {\end{definition}}

\newcommand {\bpro}[1]{\begin{property}
                      \mymarginpar{pro:#1}
                      \label{pro:#1} }
\newcommand {\epro}   {\end{property}}

\newcommand {\bprop}[1]{\begin{proposition}
                      \mymarginpar{prop:#1}
                      \label{prop:#1} }
\newcommand {\eprop}       {\end{proposition}}

\newcommand {\blem}[1]{\begin{lemma}
                      \mymarginpar{lem:#1}
                      \label{lem:#1} }
\newcommand {\elem}   {\end{lemma}}

\newcommand {\bthe}[1]{\begin{theorem}
                      \mymarginpar{the:#1}
                      \label{the:#1} }
\newcommand {\ethe}   {\end{theorem}}
\newcommand {\rthe}[1]{Theorem \ref{the:#1}}

\newcommand {\bproof}{\noindent {\bf Proof.} \ }
\newcommand {\eproof} {\hfill \squares \\ \vspace{.3cm}}
\newcommand {\bcor}[1]{\begin{corollary}
                      \mymarginpar{cor:#1}
                      \label{cor:#1} }
\newcommand {\ecor}   {\end{corollary}}

\newcommand {\bax}[1]{\begin{axiom}
                      \mymarginpar{ax:#1}
                      \label{ax:#1} }
\newcommand {\eax}       {\vspace{-.1in} \end{axiom}}

\newcommand {\bex}[2]{\vspace{.1in}
                      \begin{example}
                      \mymarginpar{ex:#1}
                       {\bf #2}
                      \label{ex:#1} }
\newcommand {\eex}       {\end{example} \vspace{.3cm} }

\newcommand {\brem}[1]{\begin{remark}
                      \mymarginpar{rem:#1}
                      \label{rem:#1} \em }
\newcommand {\erem}   {\end{remark}}

\newcommand {\beq}[1]{\mymarginpar{eq:#1}
                      \begin{equation}
                      \label{eq:#1} }

\newcommand {\beqno}[1]{\mymarginpar{eq:#1}
                      \begin{eqnarray}
                      \nonumber}

\newcommand {\eeq}       {\end{equation}}
\newcommand {\eeqno}       { && \end{eqnarray}}
\newcommand {\req}[1]{(\ref{eq:#1})}

\newcommand {\bear}[1]{\mymarginpar{eq:#1}
                       \begin{eqnarray}
                       \label{eq:#1} }

\newcommand {\bearno}[1]{\mymarginpar{eq:#1}
                       \begin{eqnarray}
                       \nonumber}

\newcommand {\eear}{\end{eqnarray}}
\newcommand {\eearno}{\end{eqnarray}}
\newcommand {\bsel}{\left \{ \begin{array}{cl}}
\newcommand {\esel}{\end{array} \right.}

\newcommand {\bmat}[1]{\left [ \begin{array}{#1}}
\newcommand {\emat}{\end{array} \right ]}
\newcommand {\bsec}[2]{\mymarginpar{sec:#2}
                       \section{#1}
                       \label{sec:#2} }

\newcommand {\bsubsec}[2]{\mymarginpar{sec:#2}
                       \subsection{#1}
                       \label{sec:#2} }

\def\R{I\kern-0.30em R}
\def\N{I\kern-0.30em N}
\def\P{I\kern-0.30em P}
\newcommand\squares{\vrule height6pt width7pt depth1pt}

\def\ex{{\bf\sf E}}
\def\pr{{\bf\sf P}}

\begin{document}
\title{A Simple Explanation for the Phase Transition in Large Language Models with List Decoding}

\author{Cheng-Shang Chang}
\orcid{0000-0002-5386-4756}
\affiliation{%
	\institution{Institute of Communications Engineering,
		National Tsing Hua University}
	\city{Hsinchu}
	\country{Taiwan, R.O.C.}}
\email{cschang@ee.nthu.edu.tw}

\begin{abstract}
Various recent experimental results show that large language models (LLM) exhibit emergent abilities that are not present in small models. System performance
is greatly improved after passing a certain critical threshold of scale. In this letter, we provide a simple explanation for such a phase transition phenomenon. For this, we model an LLM as a sequence-to-sequence random function.
Instead of using instant generation at each step, we use a list decoder that keeps a list of candidate sequences at each step and defers the generation of the output sequence at the end. We show that there is a critical threshold such that the expected number of {\em erroneous} candidate sequences remains bounded when an LLM is below the threshold, and it grows exponentially when an LLM is above the threshold. Such a threshold is related to the basic reproduction number in a contagious disease.

\end{abstract}

\maketitle

\bsec{Introduction}{introduction}

In recent years, language models have transformed natural language processing (NLP). It's now widely acknowledged that scaling up language models, such as increasing the training compute and model parameters, can enhance their performance and sample efficiency across a wide range of downstream NLP tasks (see, e.g., \cite{devlin2018bert,brown2020language}).
In many cases, scaling laws can be used to predict the impact of scale on performance \cite{kaplan2020scaling,hoffmann2022training}.
However, some downstream tasks' performance does not continuously improve with scale, which is contrary to expectations, and these tasks cannot be predicted in advance \cite{ganguli2022predictability}.

In recent papers \cite{wei2022emergent,wei2022inverse,GPT4}, it was shown by various numerical examples that
large language models (LLM) exhibit emergent abilities that are not present in small models. System performance
is greatly improved after passing a certain critical threshold of scale.
Such a phenomenon is commonly known as a {\em phase transition} in network science (see, e.g., the book \cite{Newman2010}). As indicated in \cite{wei2022emergent}, there are currently limited persuasive justifications for the manner in which these abilities develop.

The main objective of this letter is to provide a simple explanation for this phase transition phenomenon in LLMs.
For this, we model an LLM as a sequence-to-sequence random function on a certain token space with $M$ possible tokens. We assume there is an oracle that can always generate the desired output sequence to complete a requested task by the prompt sequence. The accuracy of an LLM is determined by the probability that an LLM can generate the same output sequence by the oracle. Instead of using the instant selection from a set of eligible tokens at each step in most LLMs in the literature, we use a list decoder that keeps a list of candidate sequences at each step and defers the generation of the output sequence at the end. At each step, an LLM examines a candidate sequence and enlarges it by appending a token that the LLM classifies to be eligible. There might be multiple eligible tokens for a candidate sequence, and the number of candidate sequences might grow with respect to the number of steps. We show that if the eligible token classier at each step has a bounded false alarm probability $\epsilon$ and $M \epsilon <1$, then the expected number of erroneous sequences (that are different from the oracle sequence) is bounded by a constant at each step, and thus the accuracy of an LLM can be guaranteed. On the other hand, if $M \epsilon>1$, then the expected number of erroneous sequences grows exponentially with respect to the number of steps, and there is no guarantee for accuracy. As such, there is a critical point $M \epsilon=1$. Transformer-based LLMs with more parameters and more training can memorize more patterns
\cite{vaswani2017attention,ramsauer2020hopfield} and thus are
more likely to bring down the false alarm probability $\epsilon$ below the percolation threshold $1/M$.

\bsec{Mathematical analysis}{analysis}

\bsubsec{Mathematical formulation for LLMs}{formulation}

LLMs such as GPT-4 \cite{GPT4} and PaLM-E \cite{driess2023palm} take a sequence of tokens as their input (prompts) and generate another sequence of tokens as their output (answers). To model these,
denote by ${\mathcal I}$ (resp. ${\mathcal O}$) be the input (resp. output) token space.
Without loss of generality, we
assume that both
the maximum length of an input  sequence and that of an output sequence are bounded by $N$.
An LLM can be represented by a random function
$$\phi: {\mathcal I}^{N} \mapsto {\mathcal O}^{N}.$$
Denote by ${\bf x}=(x_1, x_2, \ldots, x_N)$ (resp. ${\bf y}=(y_1, y_2, \ldots, y_N)$)  be the input (resp. output) sequence. For $1 \le t \le N$, let ${\bf y}_{1:t}=(y_1, y_2, \ldots, y_t)$.
An LLM is said to be {\em autoregressive} if $y_{t+1}$ is a random function of ${\bf x}$ and ${\bf y}_{1:t}$.
To generate $y_{t+1}$, a common practice for a Transformer-based autoregressive LLM is to generate a set of candidate tokens (e.g., the top-p sampling \cite{brown2020language}) and use a selection method (e.g., the Boltzmann selection) to select one token from the candidate set. One major drawback of an autoregressive LLM is that an incorrect selection for $y_{t+1}$ might lead to a cascaded ``hallucination'' (that generates an unrelated output). A better approach to address such a problem is to keep the candidate set in each step and defer the selection toward the end. Such an approach is commonly known as {\em list decoding} for communication systems (see, e.g., \cite{elias1957list,tal2015list,amalladinne2020coded}).

\bsubsec{An oracle}{oracle}

To compare the performance of an LLM, we assume that there is an oracle that can always generate the desired {\em deterministic} output sequence to complete the task requested by a prompt. Denote by $\phi^o$ the oracle function and ${\bf y^o}=(y_1^o, y_2^o, \ldots, y_N^o)$ be the oracle output sequence (with respect to the input $\bf x$). Similarly, let ${\bf y}^o_{1:t}=(y_1^o, y_2^o, \ldots, y_t^o)$.
The accuracy of an LLM $\phi$ (for the input $\bf x$) is defined as
\beq{accuracy1111}
\pr (\phi ({\bf x})= \phi^o ({\bf x})).
\eeq
Intuitively, an LLM with a large accuracy is more likely to complete a requested task than one with a low accuracy.

\bsubsec{Equivalent representation of an autoregressive LLM with list decoding as a sequence of binary classifiers}{classifier}

For an autoregressive LLM with list decoding, we need to keep track of candidate sets in each step.
For this, let $L_t$ be the set of candidate sequences after the $t$-th step.
The process of   generating candidates in the $t+1$-th step is equivalent to using
a binary classifier to classify all the tokens in $\mathcal O$ into two sets: {\em eligible} (with output 1) and {\em not eligible} (with output 0).
Denote such a binary classifier by the random function:
$${\mathcal C}_{t+1}(y|{\bf x},{\bf y}_{1:t})$$ for $y \in {\mathcal O}$ given a prompt {\bf x} and a sequence ${\bf y}_{1:t}$.

\begin{figure}[h]
	\centering
	\includegraphics[width=0.6\textwidth]{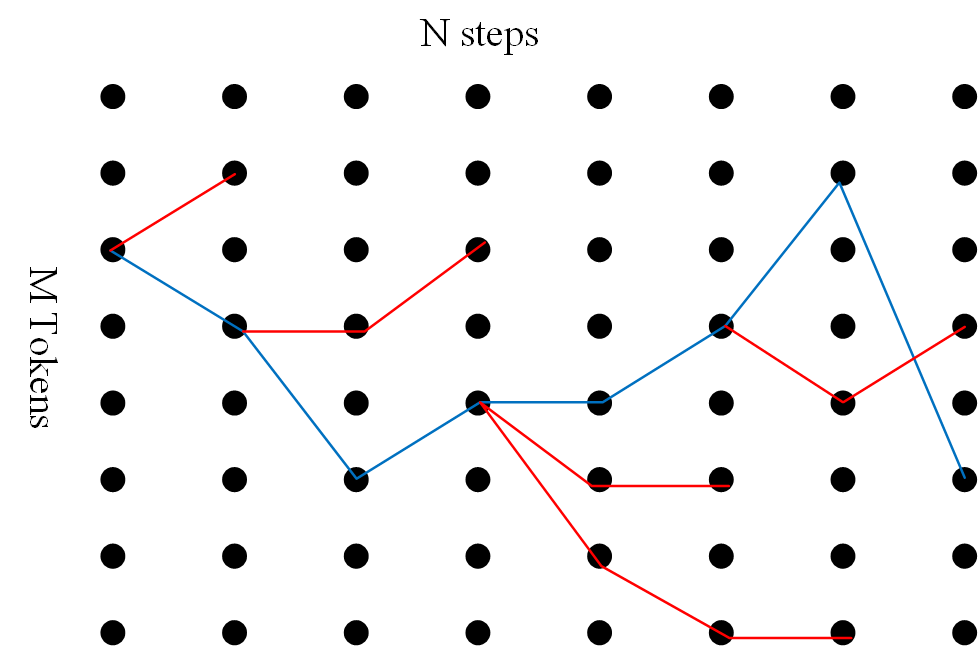}
	\caption{An illustration of the evolution of the candidate set with respect to the number of steps. The oracle sequence is marked in blue. The erroneous sequences are marked in red.}
	\label{fig:example}
\end{figure}

For a candidate sequence $ {\bf y}_{1:t} \in L_t$, if a token $y$ is eligible, i.e.,
$${\mathcal C}_{t+1}(y|{\bf x},{\bf y}_{1:t})=1,$$
then the sequence ${\bf y}_{1:t}+y$
 is  added to $L_{t+1}$, where $+$ denotes the concatenation operation.
Note that for a given ${\bf x}$ and a given ${\bf y}_{1:t}$, there might be multiple $y's$ that are classified as eligible tokens at the $t+1$-{th} step. On the other hand, it is also possible that none of them are eligible.
As such, the number of candidate sequences in $L_{t}$ might vary with respect to $t$.
In \rfig{example}, we illustrate the evolution of the candidate set with respect to the number of steps. The oracle sequence is marked in blue. The erroneous sequences are marked in red.

In the following, we define the notion of $\epsilon$-compatible LLMs that could lead to comparable performance to the oracle when $\epsilon$ is very small. The key insight of this, as shown in \rfig{example}, is that the length of an erroneous sequence cannot be too long, and it will vanish in a small number of steps.

\bdefin{compatible} An autoregressive LLM with list decoding is said to be $\epsilon$-compatible (to the oracle)
if it satisfies the following two properties:
\begin{description}
\item[(i)] 100\% recall (no missing error): the output sequence generated by the oracle is always in the candidate sets for all $t$. Specifically,
if ${\bf y}_{1:t}+y ={\bf y}^o_{1:t+1}$, then
\beq{comp1111}
{\mathcal C}_{t+1}(y|{\bf x},{\bf y}_{1:t})=1
\eeq
for all $t$ and ${\bf x}$.
\item[(ii)] bounded false alarm probability: the false alarm probability is bounded by $\epsilon$. Specifically,
if
$${\bf y}_{1:t}+y \ne {\bf y}^o_{1:t+1},$$ then
\beq{comp2222}
\pr \Big ({\mathcal C}_{t+1}(y|{\bf x},{\bf y}_{1:t})=1 \Big) \le \epsilon
\eeq
for all $t$.
\end{description}
\edefin

\bsubsec{A sufficient condition for guaranteed accuracy}{sufficient}

A sequence ${\bf y}_{1:t}$ in $L_t$ is said to be {\em erroneous} if ${\bf y}_{1:t}\ne {\bf y}^o_{1:t}$.
For an $\epsilon$-compatible LLM, we know that ${\bf y}^o_{1:t}$ is in $L_t$, and thus the number of
erroneous sequences in $L_t$ is exactly $|L_t|-1$. In the following theorem, we show that the expected number of
erroneous candidate sequences is bounded by a constant for an $\epsilon$-compatible LLM if
$M \epsilon<1$, where $M$ is the total number of tokens in the output token space ${\mathcal O}$.



\bthe{main}
Let   $R_t=|L_t|-1$ be the number of erroneous candidate sequences in $L_t$.
For an $\epsilon$-compatible LLM, if $M \epsilon <1$, then for all $t$
\beq{main0000}
 \ex [R_{t}] \le \frac{M \epsilon}{1-M \epsilon}.
 \eeq
  \ethe

\bproof
We first show that
\beq{main1111}
 \ex[R_{t+1} |L_t] \le (M \epsilon) (R_t +1).
 \eeq
 Since the LLM is $\epsilon$-compatible, we know that the sequence ${\bf y}^o_{1:t}$ must be in $L_t$.
 On average, there are (less than) $(M-1)\epsilon$ erroneous sequences of the form ${\bf y}^o_{1:t}+y$ in $L_{t+1}$.
 On the other hand, an erroneous sequence ${\bf y}_{1:t}$ in $L_t$ generates on average $M \epsilon$ erroneous sequences of the form ${\bf y}_{1:t}+y$ in $L_{t+1}$. Since there are $R_t$ erroneous sequences in $L_t$,
 the expected number of erroneous sequences in $L_{t+1}$ (given $L_t)$ is bounded above by $M \epsilon R_t+ (M-1)\epsilon$. This shows the inequality in \req{main1111}. Taking the expectation on both sides of \req{main1111} leads to
\beq{main2222}
 \ex[R_{t+1} ] \le (M \epsilon) (\ex[R_t] +1).
 \eeq
 Since $R_0=0$ and $M \epsilon <1$, it is easy to show by induction that the inequality in
 \req{main0000} holds.

\eproof

As a direct consequence of the Markov inequality, we have from \rthe{main} that
\beq{main3333}
\pr( R_t=0) =1- \pr (R_t \ge 1) \ge \frac{1-2M\epsilon}{1-M\epsilon}.
\eeq
Thus, with a nonzero probability $\frac{1-2M\epsilon}{1-M\epsilon}$, an $\epsilon$-compatible LLM can generate the same
output sequence as the oracle. The accuracy for an $\epsilon$-compatible LLM is at least $\frac{1-2M\epsilon}{1-M\epsilon}$.

\bsubsec{The phase transition}{phase}

To see the phase transition,
suppose that the inequality in \req{comp2222} is reversed and $M \epsilon>1$. Following the same induction argument in the proof of \rthe{main}, one can show that
\beq{main4444}
\ex[R_t] \ge (M \epsilon)^{t}.
\eeq
Thus,
the expected number of erroneous sequences in $L_t$ grows exponentially with respect to the number of steps. As such,
the accuracy of the LLM is low.

In view of \req{main0000} and \req{main4444}, the critical point for the phase transition is $M \epsilon =1$.
This role of $M \epsilon$ is analogous to the basic reproduction number in a contagious disease \cite{Newman2010,chen2020time}. If the basic reproduction number is smaller than 1, then the disease can be contained. On the other hand, if the basic reproduction number is larger than 1, then it is likely to have a large outbreak.

\bsec{Conclusion}{con}

In this letter, we provided a simple explanation for the phase transition in LLMs. For an $\epsilon$-compatible LLM, we showed that the expected number of erroneous sequences is bounded by a constant and the accuracy of the LLM can be guaranteed when $M \epsilon <1$. On the other hand, if  the inequality in \req{comp2222} is reversed and $M \epsilon>1$, then the expected number of erroneous sequences grows exponentially with respect to the number of steps.

One possible extension of the list decoder is to track the probability of each candidate sequence. For most Transformer-based LLMs, it is possible to compute such a probability. Then the output sequence is generated by the candidate sequence with the largest probability at the end.


\begin{acks}

This work was supported in part by the National Science and Technology under Grant
MOST 111-2221-E-007-045-MY3.
\end{acks}

\end{document}